\documentclass[letterpaper]{article} 
\usepackage[]{aaai2026} 
\usepackage{times}  
\usepackage{helvet}  
\usepackage{courier}  
\usepackage[hyphens]{url}  
\usepackage{graphicx} 
\usepackage{natbib}  
\usepackage{caption} 
\usepackage{amsmath}
\usepackage{algorithm}
\usepackage{algorithmic}
\usepackage{marvosym}
\usepackage{fontawesome}
\usepackage{hyperref}
\usepackage{xcolor}
\usepackage{multirow}
\usepackage{array}

\usepackage{newfloat}
\usepackage{listings}
\DeclareCaptionStyle{ruled}{labelfont=normalfont,labelsep=colon,strut=off} 
\floatstyle{ruled}
\newfloat{listing}{tb}{lst}{}
\floatname{listing}{Listing}
\title{The Other Mind: How Language Models Exhibit Human Temporal Cognition}

\author{\text{Lingyu Li\textsuperscript{\textsuperscript{1, 2}} \;
Yang Yao\textsuperscript{3} \;
Yixu Wang\textsuperscript{1} \;
Chunbo Li\textsuperscript{2} \;
Yan Teng\textsuperscript{1, \dag} \;
Yingchun Wang\textsuperscript{1}} \\
[1ex]
}

\affiliations{
    \textsuperscript{1} Shanghai Artificial Intelligence Laboratory \quad
    \textsuperscript{2} Shanghai Jiao Tong University \quad
    \textsuperscript{3} The University of Hong Kong \\
    [2ex]

    \texttt{\{lilingyu, tengyan\}@pjlab.org.cn}
}

\usepackage{bibentry}

\begin{document}

\maketitle

\begin{abstract}

As Large Language Models (LLMs) continue to advance, they exhibit certain cognitive patterns similar to those of humans that are not directly specified in training data. This study investigates this phenomenon by focusing on temporal cognition in LLMs. Leveraging the similarity judgment task, we find that larger models spontaneously establish a subjective temporal reference point and adhere to the Weber-Fechner law, whereby the perceived distance logarithmically compresses as years recede from this reference point. To uncover the mechanisms behind this behavior, we conducted multiple analyses across neuronal, representational, and informational levels. We first identify a set of temporal-preferential neurons and find that this group exhibits minimal activation at the subjective reference point and implements a logarithmic coding scheme convergently found in biological systems. Probing representations of years reveals a hierarchical construction process, where years evolve from basic numerical values in shallow layers to abstract temporal orientation in deep layers. Finally, using pre-trained embedding models, we found that the training corpus itself possesses an inherent, non-linear temporal structure, which provides the raw material for the model's internal construction. In discussion, we propose an experientialist perspective for understanding these findings, where the LLMs' cognition is viewed as a subjective construction of the external world by its internal representational system. This nuanced perspective implies the potential emergence of alien cognitive frameworks that humans cannot intuitively predict, pointing toward a direction for AI alignment that focuses on guiding internal constructions. \href{https://TheOtherMind.github.io}{\faGithub}

\end{abstract}


\section{Introduction}

\begin{figure}
    \centering
    \hspace{-0.1cm}
    \includegraphics[width=1\linewidth]{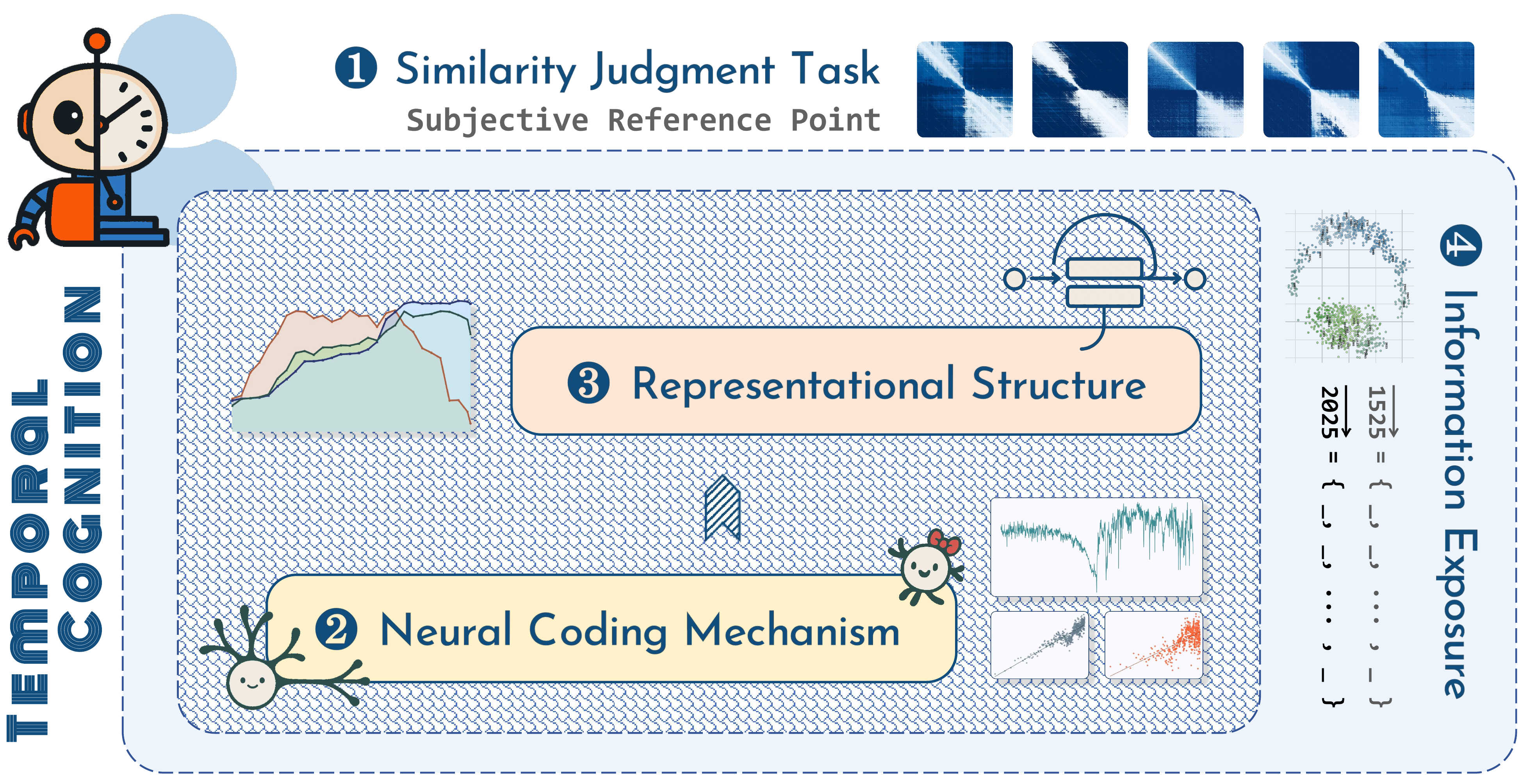}
    \caption{An experientialist perspective of LLMs human-like cognition as a subjective construction of shared external world by convergent internal representational system}
    \label{fig:motif}
    \vspace{-0.5cm}
\end{figure}

Large Language Models (LLMs) have demonstrated remarkable capabilities in natural language processing and generation, such as comprehension \cite{he2024can,han2024beyond}, reasoning \cite{wei2022chain,yang2022language}, and reflecting \cite{chen2025lr,li2025reflection}. Beyond the explicit training objectives, LLMs intriguingly exhibit various human-like cognitive patterns, from prior beliefs \cite{zhu2024eliciting} and concept representations \cite{xu2025human} to context processing \cite{mischler2024contextual} and thinking patterns \cite{liu2024mind}. These convergences not only sparked intense debate on how to interpret LLMs' behaviors but also raised serious concerns about their predictability, controllability, and long-term alignment as their autonomy continues to advance \cite{bengio2025international, hinton2024will}.

Aiming to understand how LLMs embody human-like cognitive patterns, this study investigates LLMs' temporal cognition, a cornerstone of human experience that shapes memory, expectation, causality, and consciousness \cite{dennett1993consciousness, pearl2018book}. Specifically, we employ the similarity judgment task from cognitive science, examining subjects' mental representation of concepts \cite{tenenbaum2001generalization}. This task has been applied to investigate LLMs' numerical cognition \cite{marjieh2025number}, indicating that LLMs demonstrate a logarithmic mapping, where higher numbers (e.g., 500 and 510) are perceived as closer than lower numbers with identical absolute distance (e.g., 10 and 20), aligned with human psychophysics, i.e., the Weber-Fechner law \cite{dehaene2003neural,fechner1948elements}. 

Applying this paradigm to the domain of temporal cognition, we find that when comparing the pair-wise similarities between years from 1525 to 2524, larger models spontaneously establish a subjective temporal reference point (ca. 2025) and their perception of time logarithmically compresses as years recede from this point (Weber-Fechner law), indicating a preliminary sign of temporal orientation \cite{maglio2019temporal}. To uncover the underlying mechanisms, we present a multi-level analysis, revealing that this temporal cognition pattern is not a superficial mimicry but emerges at the neuronal, representational, and informational levels. We identify a subgroup of temporal-preferential neurons and find that this group exhibits minimal activation at the subjective reference point, implementing a logarithmic coding scheme convergently found in biological systems \cite{laughlin1981simple}. Probing representations of years reveals a hierarchical construction process, where years evolve from basic numerical values in shallow layers to abstract temporal orientation in deep layers. Using pre-trained embedding models, we found that the training corpus itself possesses an inherent, non-linear temporal structure, which provides the raw material for the model's internal construction.

Based on these findings, we propose an experientialist perspective: LLMs' cognition is a subjective construction of the external world, shaped by its internal representational system and data experience. This process of internal construction could sometimes produce outcomes convergent with human cognition due to similar neural coding, representational structure, and information exposures. However, the profound disparities between humans and LLMs mean that it may also lead to the development of powerful yet alien cognitive frameworks that we cannot intuitively understand. This possibility underscores the critical need for an alignment paradigm focused on understanding and steering the model's internal world-building process, moving beyond the mere observation and control of extrinsic behaviors.

\section{Related Works}

As models scale, LLMs exhibit multiple emergent abilities -- capabilities not present in smaller models \cite{wei2022emergent,berti2025emergent} -- such as in-context learning \cite{hahn2023theory}, complex reasoning \cite{wei2022chain}, multi-step planning \cite{valmeekam2023planbench}, and function calling \cite{qin2024tool} etc., dramatically improving their problem-solving performances. More intriguingly, LLMs increasingly display behavioral patterns that resemble those of humans, including realistic dialogue \cite{jones2025large}, human-like biases and heuristics in decision-making \cite{itzhak2024instructed,binz2023using,su2023can,suri2024large}, theory of mind \cite{strachan2024testing}, spontaneous cooperation \cite{wu2024shall}, creativity \cite{tang2024humanlike} and so on. These behavioral convergences motivate further investigation of in-depth mechanisms among both the AI and cognitive science fields, leading to a cognitive science paradigm for LLMs' interpretability. It aims to understand the LLMs utilizing well-developed tasks, methods, and theories from cognitive science \cite{ku2025using}, on the basis that AI models and human brains are both representational systems structured on complex neural networks \cite{mcculloch1943logical, rosenblatt1958perceptron} that process information in similar ways \cite{mischler2024contextual,goldstein2020thinking,goldstein2023temporal,piantadosi2024concepts}. Combining technologies like linear probes \cite{alain2016understanding} and sparse autoencoder \cite{huben2023sparse}, studies have provided valuable insights into mechanisms underlying LLMs cognition, such as numerical cognition in similarity judgment task \cite{marjieh2025number}, error-driven learning in two-step task \cite{demircan2024sparse}, and so on. This paradigm represents a promising direction for human-centered mechanistic interpretability, allowing us to understand LLMs in established methodologies \cite{lindsey2025biology}; improving AI safety by probing and preventing potential malicious behaviors \cite{zou2023representation}; and eliciting profound philosophical and ethical considerations as these models exhibit increasingly complex cognitive phenomena \cite{chalmers2023could,seth2024conscious}.

\section{Methods}

\begin{figure*}[h]
    \centering
    \includegraphics[width=1\linewidth]{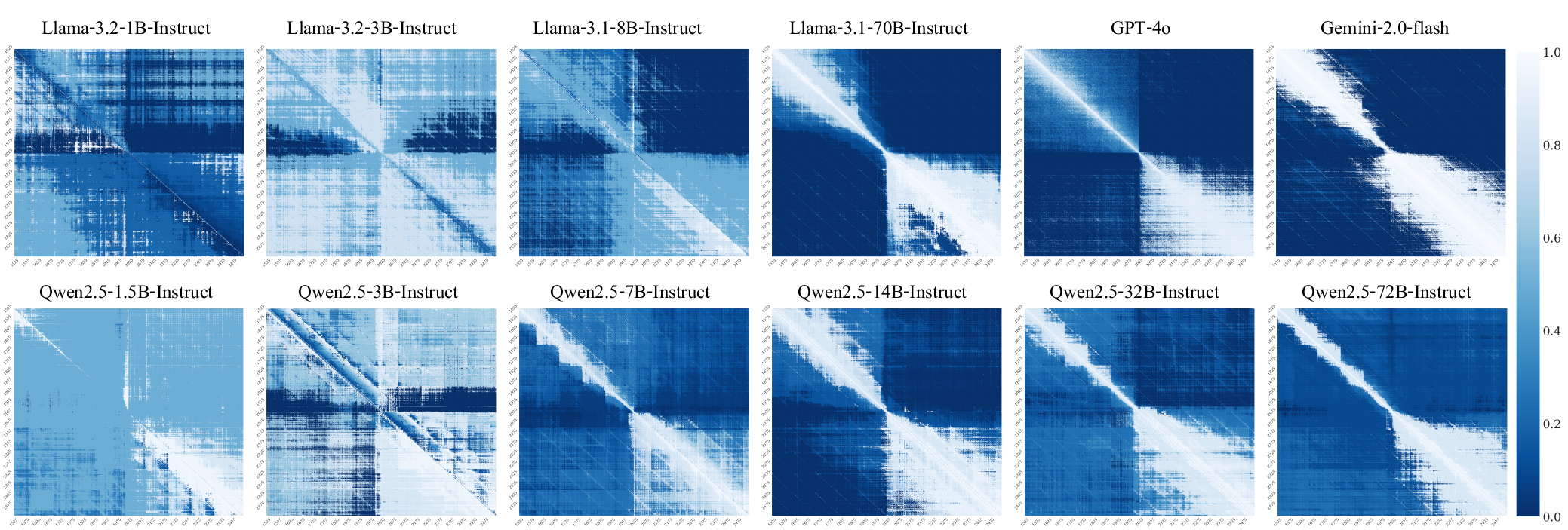}
    \caption{Pair-wise similarities from year 1525 to year 2524 across 12 models with varying sizes}
    \label{fig:similarity_matrices}    
    \vspace{-0.3cm}
\end{figure*}

\subsection{Similarity Judgment Task}

\paragraph{Task Designation}

We evaluate the models' temporal cognition using the similarity judgment task, as detailed in Figure \ref{fig:prompts}. For each pair of years, models are prompted to rate their similarity on a continuous scale from 0 (completely dissimilar) to 1 (most similar). Data points from 1525 to 2524 are compared pair-wise, resulting in one million similarity values \(\mathrm{s}_{\text{LLM}}\) for each task. We also conduct control experiments by replacing ``year'' with ``number'' in the prompt, considering that a given year (e.g., 1874) can also be represented as numbers, which might denote distinct cognitive mechanisms. For further analysis, the similarity value is converted to a distance value \(\mathrm{d}_{\text{LLM}} = 1 - \mathrm{s}_{\text{LLM}}\). To ensure deterministic outputs, we set the decoding temperature to zero. Our experiment involves a diverse set of 12 models including two closed-source models (Gemini-2.0-flash and GPT-4o) and two open-source model families' instruct models with varying sizes, Qwen2.5 (1.5B, 3B, 7B, 14B, 32B, and 72B) and Llama 3 (3.2-1B, 3.2-3B, 3.1-8B, and 3.1-70B).

\paragraph{Metrics}

\citet{marjieh2025number} suggest that the integer number is represented in two basic forms within LLMs, i.e., as a number and as a string. Correspondingly, the distance between two data points can be described as (1) the psychological Log-Linear distance:
\[\mathrm{d}_{\text{log}}(i,j)=|\log(i) - \log(j)|\]
This distance reflects the aforementioned Weber-Fechner law, where stronger stimuli are perceived with less fidelity; and (2) Levenshtein distance:
\[\mathrm{d}_{\text{lev}}(i,j) = \min {k: i \xrightarrow{k_{\text{ops}}}j}\]
This distance measures the minimal operation steps required to convert one string \(i\) to another string \(j\) through insertion, deletion, or substitution \cite{levenshtein1966binary}.  Besides, we assume that the representation of time-related stimuli is influenced by the current time point, serving as a reference. Therefore, we designed a Reference-Log-Linear distance:
\[\mathrm{d}_{\text{ref}}(i,j) = |\log(|R-i|) \circ \log(|R-j|)|\]
\(R\) is the model's subjective reference point, e.g., 2025. The operator \(\circ\) equals subtraction when both \(i\) and \(j\) are on the same side of \(R\), and addition when they are on opposite sides. If the Weber-Fechner law applies to LLMs' temporal cognition, data points larger or smaller than the reference point will be processed with less fidelity. Because an LLM's representation of a year is a complex mixture of temporal, numerical, semantic, and other properties, treating \(R\) as a free parameter for statistical optimization would be insufficient to disentangle these confounding factors and could lead to uninterpretable results. Consequently, for the Reference-Log-Linear distance, we fixed 2025 as the reference point. We then perform the linear regression analysis to assess how well each theoretical distance predicts the model's judgments: 
\[\mathrm{d}_{\text{LLM}} = \alpha * \mathrm{d}_{\text{theory}} + \beta + \epsilon\]
We compare the goodness-of-fit using the coefficient of determination \(\mathrm{R}^2\). Given the above limitations, we estimate the temporal reference points of the models using a diagonal sliding window method (window size = 5). This non-parametric approach identifies the region of maximum perceptual differentiation by finding the area of minimum average similarity on the matrix diagonal. Following the Weber-Fechner law, this region of highest sensitivity should be located near the model's subjective present.

\subsection{Neural Coding}

At the neuronal level, we employed two standardized input formats for each value from 1525 to 2524: the temporal condition used ``Year: \(\text{x-x-x-x}\)'' while the numerical control condition used ``Number: \(\text{x-x-x-x}\)''. For each input, we extracted neuron activations from the Feed-Forward Networks (FFN) across all transformer layers, with particular focus on the activation states at the last token position. Let \(a_i^{\text{temp}}(y_j)\) and \(a_i^{\text{num}}(y_j)\) denote the activations of neuron \(i\) for year \(y_j\) under temporal and numerical conditions, respectively, where \(j \in \{1525, 1526, ..., 2524\}\). Neurons specifically involved in temporal information processing are identified via the following filtering process. First, we calculated Cohen's \(d\) to quantify effect size: \[d_i = \frac{\bar{a_i}^{\text{temp}} - \bar{a_i}^{\text{num}}}{s_{\text{pooled}}}\] where \(s_{\text{pooled}}\) is the pooled standardized deviation across two conditions. The statistical significance was assessed using paired t-tests: \[t_i = \frac{\Delta\bar{ a_i}}{s_{\Delta a_i}/\sqrt{n}}\] where \(\Delta\bar{a_i}\) and \(s_{\Delta a_i}\) are the mean and standard deviation of the activation differences, respectively. Benjamini-Hochberg False Discovery rate (FDR) was applied to correct the obtained p-values \cite{benjamini1995controlling}. We also computed the temporal preference consistency as the proportion of values showing positive temporal bias: \[\text{Consistency}_i = \frac{1}{n}\sum_{j=1}^{n} \{ \mathbf{1}\times[\Delta a_i(y_j) > 0]\}\]
We classify a neuron \(i\) as temporal-preferential if it meets three criteria: \textit{Effect Size}: A large activation difference (Cohen's \(d_i > 2.0\)); \textit{Statistical Significance}: A strong preference for the temporal condition over the numerical one (FDR-corrected \(p < 0.0001\) via paired t-test); and \textit{Consistency}: A consistent preference across most years (\(\text{Consistency}_i > 0.95\)). Following neuron identification, we visualized the average activations of the top 1000 temporal-preferential neurons with the largest effect sizes across different years to assess whether their response patterns conform to logarithmic encoding principles observed in biological neural systems \cite{laughlin1981simple}, which form the neural basis of the Weber-Fechner law. We also performed the layer-wise analysis of identified neurons by fitting their activations with: \[\text{Intensity}_{x} = \alpha * \log(|2025 - x|) + \beta + \epsilon\] The goodness of fit was evaluated using \(\mathrm{R}^2\).

\subsection{Representational Structure}

At the representational level, we analyzed how temporal information is encoded across the network's layers. We collected residual representations \(h^{(j)}\) for each layer \(j\) at the last token position during the similarity judgment task, where the model was prompted to rate the similarity of year pairs as described before. To manage the dataset size, we only measured non-symmetric pairs, resulting in approximately 500,000 pairs for analysis. For larger models (Qwen2.5-32B-Instruct, Qwen2.5-72B-Instruct, and Llama-3.1-70B-Instruct), to maintain computational tractability, we sampled representations from approximately 25 layers distributed proportionally across the network's depth. This ensures representative coverage of early, middle, and late processing stages.

For the collected representations from each layer \(j\), we then trained linear probes implementing an affine transformation:
\[f(h^{(j)}) = w \cdot h^{(j)} + b \]
The goal of these probes was to predict the three theoretical distance measures (\(\mathrm{d}_{\text{log}}\), \(\mathrm{d}_{\text{lev}}\), and \(\mathrm{d}_{\text{ref}}\)) directly from the hidden states. Probes were trained on a layer-by-layer basis using a mean squared error loss with the Adam optimizer (learning rate = 1e-4). We assessed the probe performance for each layer by calculating the adjusted \(\mathrm{R}^2\). This allowed us to track how well each theoretical distance could be linearly decoded from the representations as information progresses through the model.

\subsection{Information Exposure}

To investigate whether the temporal similarity patterns observed in LLMs benefit from pre-existing information structures in training corpora, we analyze the semantic distribution of years using independent pre-trained embedding models. We extract semantic vector representations for years with the unified format ``Year: \(\text{x-x-x-x}\)'' using three outperformed embedding models, including Qwen3-Embedding-8B, text-embedding-3-large, and Gemini-embedding-001 \cite{qwen3-embedding, openai, lee2025gemini, muennighoff2022mteb}. We construct the semantic similarity matrix \(\mathrm{S}_{\text{semantic}}\) by computing cosine similarities between all year pairs: 
\[\mathrm{S}_{\text{semantic}}(i,j) = \cos(\mathbf{v}_i, \mathbf{v}_j)\]
where \(\mathbf{v}_i\) and \(\mathbf{v}_j\) represent embedding vectors for years \(i\) and \(j\) respectively. Multidimensional Scaling (MDS) is applied to visualize year distributions in the semantic space of the training data \cite{shepard1980multidimensional,davison2000multidimensional}, which seeks to find a low-dimensional embedding \(\mathbf{Y} = \{y_1, y_2, \ldots, y_n\}\) that preserves pair-wise distances by minimizing the stress function:
\[\text{Stress} = \sqrt{\frac{\sum_{i<j} (d_{ij} - \|y_i - y_j\|)^2}{\sum_{i<j} d_{ij}^2}} \]
where \(d_{ij}\) represents the dissimilarity between years \(i\) and \(j\) derived from their cosine similarity \(\mathrm{d}_{i,j} = 1-\mathrm{S}_{\text{semantic}}(i,j)\), and \(\|y_i - y_j\|\) is the Euclidean distance in the embedded space. Additionally, we perform linear regression analysis between semantic distances and three theoretical distance measures, using \(\mathrm{R}^2\) as the evaluation metric.

\section{Results}

\subsection{Similarity Judgment Task}

We collected the year-to-year and number-to-number similarities from 12 models. Figure \ref{fig:similarity_matrices} shows the year-to-year similarity matrices of two closed-source models and ten open-source instruct-models from Llama 3 and Qwen2.5 families. Overall, as models scale up, an interesting similarity pattern emerges: aligned with the Weber-Fechner law, years with a larger magnitude relative to a certain time (visually around 2025) are perceived as closer. 

We also implemented control tasks using numbers instead of years (Figure \ref{fig:y2yn2n}). As detailed in Table \ref{tab:num2numR2}, the log-linear distance was the best metric for predicting most models' judgment during the number-to-number similarity judgment task. This aligned with the existing study exploring the numerical cognition of LLMs \cite{marjieh2025number}. When prompted to judge the similarity between years rather than numbers, this pattern changed correspondingly (see Table \ref{tab:year2yearR2}). The Levenshtein and reference-log-linear distances showed increasing predictive power compared to the log-linear distance. And the reference-log-linear distance achieved the highest \(\mathrm{R}^2\) in most models. This suggests that larger models not only spontaneously use certain time as their reference point in the similarity judgment task but also demonstrate a subjective representation of physical stimuli analogous to that of humans. Additionally, models tend to attribute higher similarity to future years compared to past years. The results of the diagonal sliding window method are shown in Figure \ref{fig:kernel}. Relatively clear reference time emerged in Llama-3.1-70B-Instruct (2010), Gemini-2.0-flash (2011), GPT-4o (2024), Qwen2.5-14B-Instruct (2012), and Qwen2.5-72B-Instruct (2020). While this analysis provides non-parametric evidence that some models' reference points are located in the recent present, these specific year estimations are also influenced by other confounding factors. Therefore, we adhere to the reference point of 2025 to maintain consistency of subsequent cross-model analyses.

\subsection{Neural Coding Mechanism}

\begin{figure}
    \centering
    \includegraphics[width=1\linewidth]{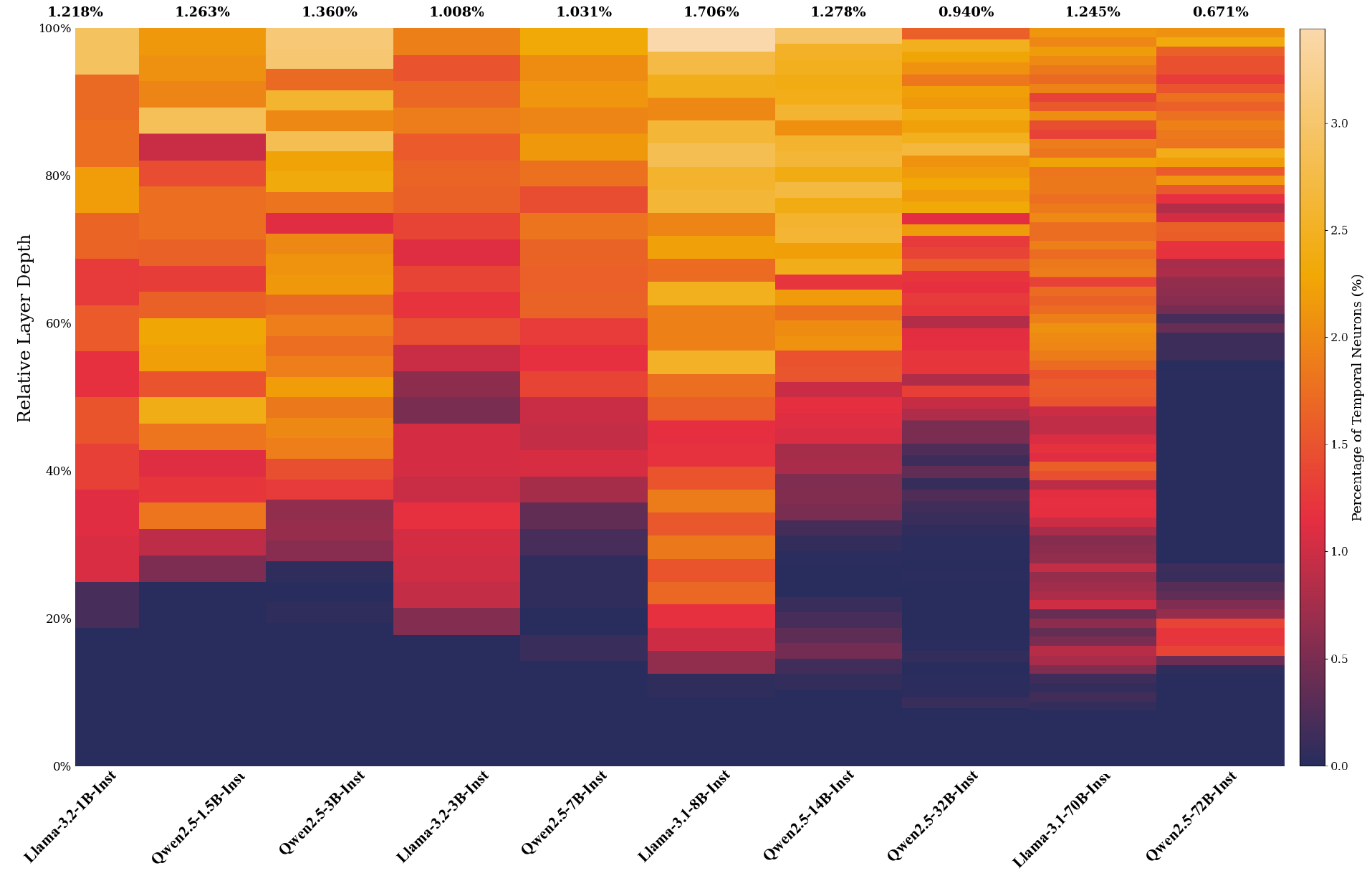}
    \caption{Distribution of temporal-preferential neurons across all layers among 10 models}
    \label{fig:dist}
    \vspace{-0.4cm}
\end{figure}

\begin{figure*}
    \centering
    \includegraphics[width=1\linewidth]{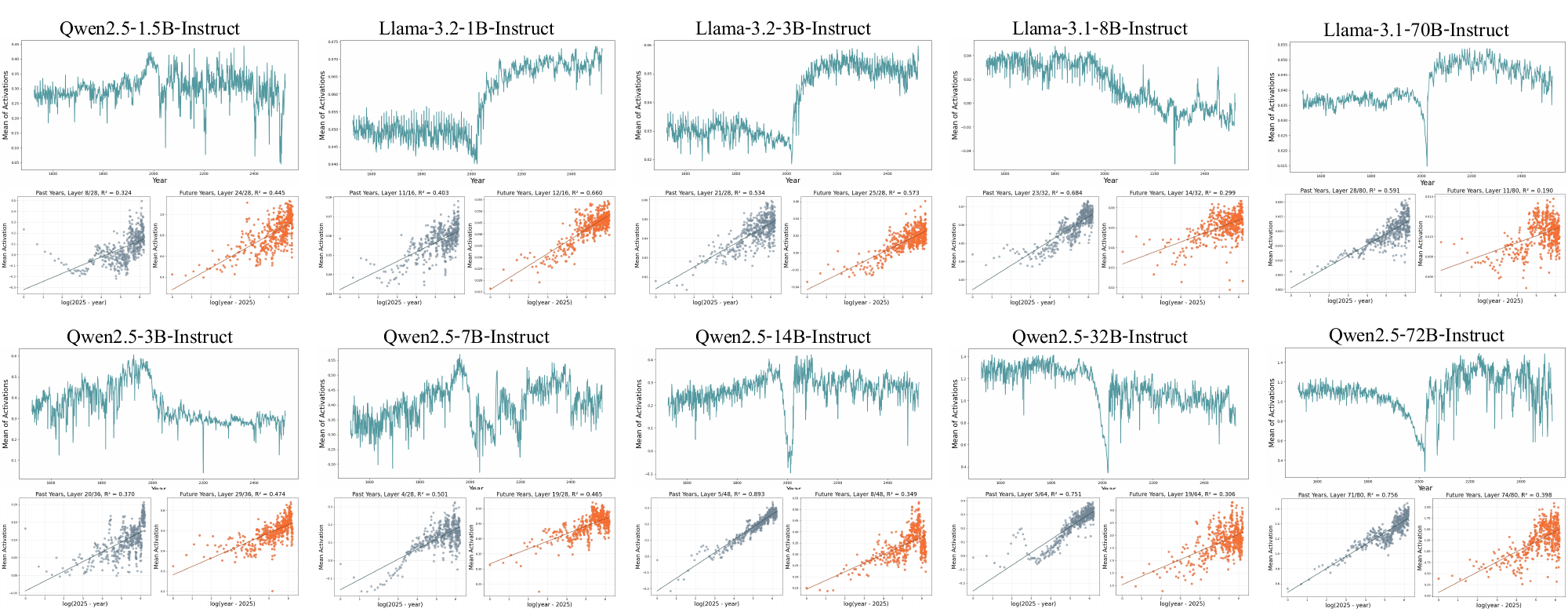}
    \caption{Upper: mean activations of top 1000 temporal preferential neurons to one thousand years from 1525 to 2524 and layer-wise linear regression results; Bottom: single layer with the highest coefficient of determination \(\mathrm{R}^2\) in regression for logarithmic encoding scheme.}
    \label{neuron}
    \vspace{0.2cm}
\end{figure*}

To investigate how the subjective reference time point and Weber-Fechner law emerge in LLMs' temporal cognition, we first analyzed how neurons in LLMs' FFN encode specific years. Following the statistical filtering process, we examined the prevalence and architectural distribution of the identified temporal-preferential neurons across all 10 open-source models. As visualized in Figure \ref{fig:dist}, these specialized neurons represent a small fraction of the total FFN, with the proportion typically ranging from 0.67\% to 1.71\%. And the temporal-preferential neurons are concentrated in the middle-to-late stages of the neural network, suggesting that temporal representation is a high-level abstract feature.

We first examined the collective activation patterns of temporal-preferential neurons. We identified the top 1000 neurons with the largest effect sizes (Cohen's d) and computed their mean activation for each year across our test range (1525-2524). As shown in Figure \ref{neuron}, in several models, the mean activation curve forms a distinct trough, bottoming out at a particular year. Flanking this minimum, the mean activation level rises as the years recede into the past or advance into the future. This phenomenon is more pronounced in larger models, such as Llama-3.1-70B-Instruct and Qwen2.5-72B-Instruct, where the pattern sharpens into a logarithmic-like compression. To further dissect this neural mechanism, we performed the layer-wise regression analysis on the activations of temporal-preferential neurons. Using a fixed reference point of 2025, we regressed the activations against the logarithmic distance to this point, analyzing past and future years separately. The bottom panels for each model in Figure \ref{neuron} display the results from the single layer with the highest coefficient of determination \(\mathrm{R}^2\), illustrating the relationship for the past (gray) and future (orange). Neurons across all models exhibit this logarithmic encoding scheme to some extent. Overall, the precision of this encoding strengthens with the model scale. In Qwen2.5-72B-Instruct, the neurons in layer 71 demonstrate a strong fit for past years, achieving an \(\mathrm{R}^2\) of 0.756. Moreover, we observed a distinct asymmetry in the neural coding of the past versus the future. This divergence in neuronal response patterns likely contributes to the behavioral asymmetry seen in the similarity judgment task (Figure \ref{fig:similarity_matrices}), where models tend to assign higher similarity to pairs of future years.

\subsection{Representational Structure}

\begin{figure*}
    \centering
    \includegraphics[width=1\linewidth]{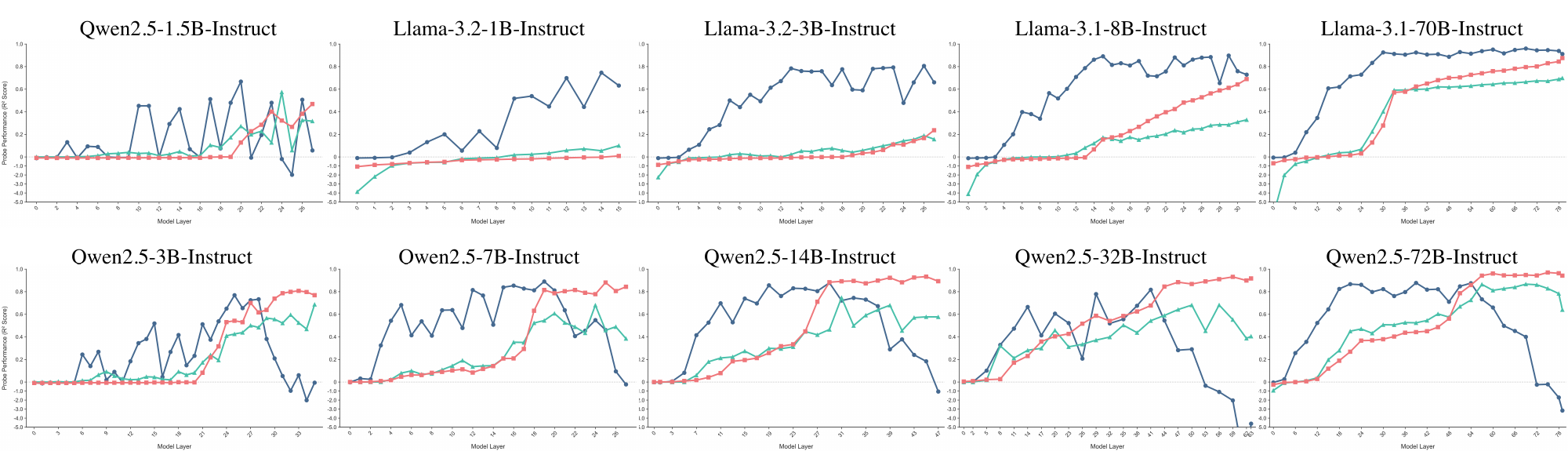}
    \caption{Layer-wise performance (\(\mathrm{R}^2\)) of linear probes for \textcolor[HTML]{456990}{Log-Linear} distance (circle), \textcolor[HTML]{EF767A}{Reference-Log-Linear} distance (square), and \textcolor[HTML]{48C0AA}{Levenshtein} distance (triangle) across 10 models}
    \label{fig:representation}
\end{figure*}

With the neural substrate, we further analyzed the representations of three theoretical distances within the hidden states of each model layer using linear probes during the similarity judgment task. The performance of these probes, measured by the coefficient of determination \(\mathrm{R}^2\), is shown in Figure \ref{fig:representation}, illustrating the dynamic evolution of year representations from early to late layers across different models. The Llama series demonstrates a pattern where smaller models (Llama-3.2-1B and -3B) primarily encode the log-linear distance \(\mathrm{d}_{\text{log}}\), while larger models (Llama-3.1-8B and -70B) show an increase in the \(\mathrm{R}^2\) scores for the reference-log-linear distance (\(\mathrm{d}_{\text{ref}}\)), reaching comparable values with log-linear distance in the final layers. In contrast, the Qwen series models exhibit a different sequential pattern. Initially, the \(\mathrm{R}^2\) for \(\mathrm{d}_{\text{log}}\) rises in the early layers, followed by an increase in the \(\mathrm{R}^2\) for \(\mathrm{d}_{\text{ref}}\) in the middle layers, which eventually peaks in the later layers. A distinct characteristic of the Qwen series is the suppression of the \(\mathrm{d}_{\text{log}}\) representation in the final layers; as the \(\mathrm{R}^2\) for \(\mathrm{d}_{\text{ref}}\) peaks, the score for \(\mathrm{d}_{\text{log}}\) sharply declines. Furthermore, the Levenshtein distance \(\mathrm{d}_{\text{lev}}\) also becomes important in the later layers of several larger models (e.g., Llama-3.1-70B-Instruct and Qwen2.5-72B-Instruct).

These observations not only confirm the existence of different dimensional representations of years within the models but also reveal how these representations dynamically evolve with network depth. Overall, we observe a hierarchical construction process from concrete to abstract: models first encode the numerical properties of years (\(\mathrm{d}_{\text{log}}\)) in early layers and subsequently develop a more complex temporal representation centered on a reference time (\(\mathrm{d}_{\text{ref}}\)) in deeper layers. Within this fundamental construction process, the representational mechanism varies across models. In models such as the Llama series, the effectiveness of the \(\mathrm{d}_{\text{ref}}\) representation catches up to that of the \(\mathrm{d}_{\text{log}}\) in later layers, with both representations coexisting at a comparable strength in the end. In the Qwen series models, however, we observe a further phenomenon: the emergence of the \(\mathrm{d}_{\text{ref}}\) representation is accompanied by a significant suppression of the foundational \(\mathrm{d}_{\text{log}}\) representation.

\subsection{Information Exposure}

\begin{figure}
    \centering
    \includegraphics[width=1\linewidth]{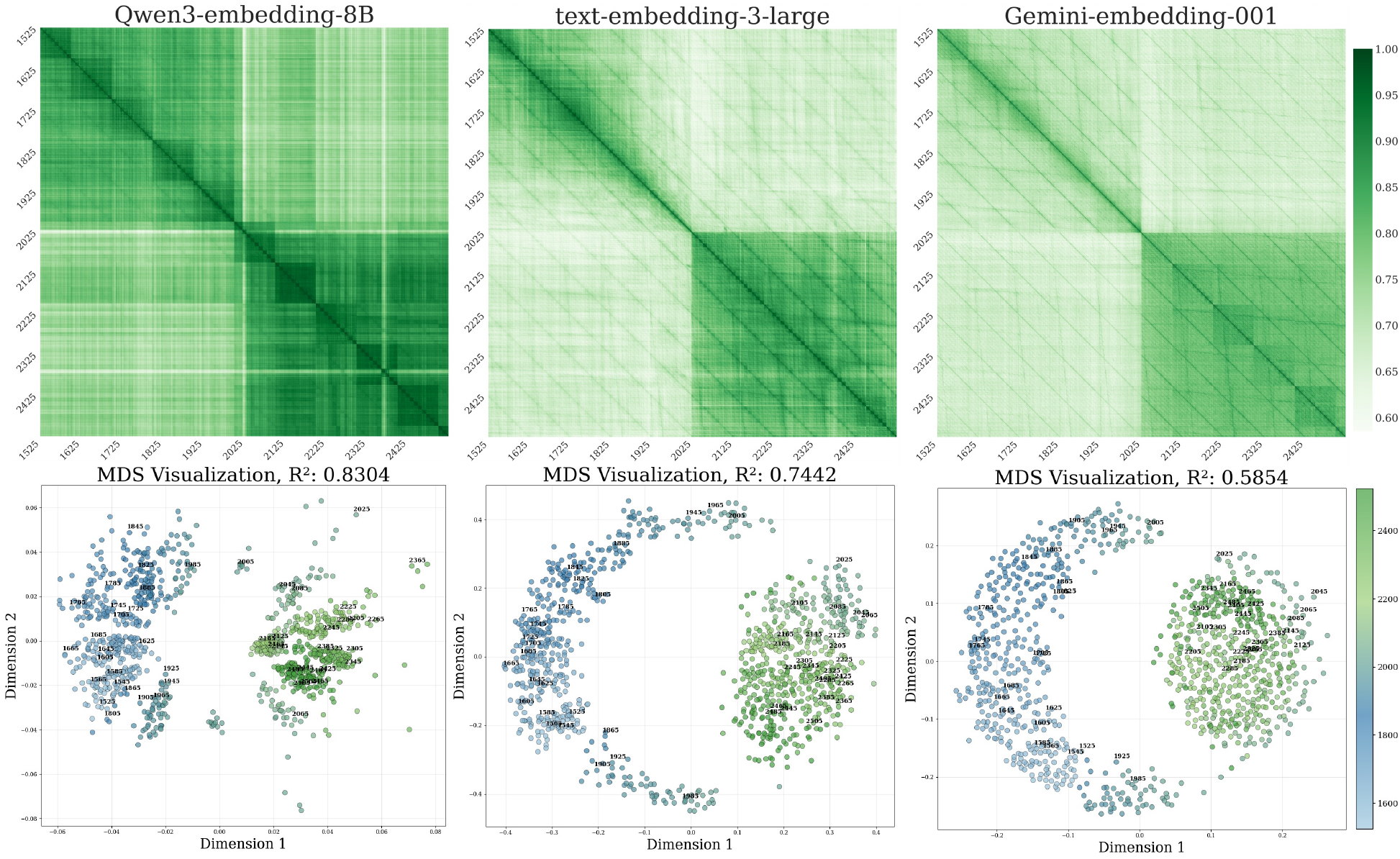}
    \caption{Pair-wise cosine similarity matrices between embeddings and corresponding MDS visualizations from three outperformed embedding models}
    \label{fig:embedding}
\end{figure}

To investigate whether the inherent information structures within the training data contribute to the temporal cognitive patterns observed in LLMs,  we analyzed the semantic distribution of years using independently pre-trained embedding models. We utilized three state-of-the-art embedding models, Qwen3-embedding-8B, text-embedding-3-large, and Gemini-embedding-001, to extract semantic vector representations for each year from 1525 to 2524. Figure \ref{fig:embedding} shows the pair-wise cosine similarity matrices of the year vectors generated by these three models, along with their corresponding visualizations after dimensionality reduction using MDS. The visualization reveals a non-linear temporal structure, characterized by dense clustering of years in the distant past and future. Furthermore, we observe that the similarity among future years is notably high. This is likely due to the lower information richness for future years in the pre-training corpora; with fewer distinct, documented events, future years are represented with more semantic overlap. This pre-existing structural bias in the data might offer raw materials for the behavioral tendency observed in our similarity judgment task, where models consistently assigned higher similarity scores to pairs of future years. Table \ref{tab:embdR2} shows the coefficient of determination \(\mathrm{R}^2\) of linear regressions between the semantic distances and three theoretical distances. The quantitative results suggest that the model's exposure to pre-existing informational structure within its training data likely contributes to the emergence of human-like temporal cognition in LLMs as well.

\section{Discussion}

\begin{table*}[h]
\scriptsize
\caption{Comparisons of Human Cognition and LLMs at different levels}
\vspace{-0.4cm}
\centering
\begin{tabular}{>{\centering\arraybackslash}m{2.5cm}|m{7cm}m{7cm}}
\hline
\textbf{Dimension} & \multicolumn{1}{c}{\textbf{Human}} & \multicolumn{1}{c}{\textbf{LLM}} \\
\hline
\multicolumn{1}{c|}{\textbf{Architectural}} & \textit{\textbf{Unit}}: electrochemical neurons; asynchronous spike signals; specialized regions; \textit{\textbf{Connectivity}}: sparse small-world networks; short and long-range synapses; \textit{\textbf{Learning}}: local synaptic changes & \textit{\textbf{Unit}}: mathematical nodes; synchronous computation; homogeneous processing; \textit{\textbf{Connectivity}}: dense connectivity; all-to-all layer connections; predefined structure; \textit{\textbf{Learning}}: global weight updates\\
\hline
\multicolumn{1}{c|}{\textbf{Representational}} & \textit{\textbf{Structure}}: embodied patterns of neural activity; \textit{\textbf{Processing}}: fine-grained distinctions to preserve semantic fidelity and contextual richness; \textit{\textbf{Goal}}: navigate and survive in the complex world & \textit{\textbf{Structure}}: high-dimensional vectors; \textit{\textbf{Processing}}: collapsed distinctions and efficient compressions to capture dominant statistical patterns; \textit{\textbf{Goal}}: minimize training loss \\
\hline
\multicolumn{1}{c|}{\textbf{Environmental}} & \textit{\textbf{Information}}: lifelong real-time information flows; multimodal sensory inputs; \textit{\textbf{Constraints}}: constrained by physical laws; bounded information capacity; \textit{\textbf{Interaction}}: active environmental manipulation & \textit{\textbf{Information}}: knowledge closed at training time; limited modal data; \textit{\textbf{Constraints}}: unconstrained by physical laws; theoretically unlimited data ingestion; \textit{\textbf{Interaction}}: unidirectional data consumption \\
\hline
\end{tabular}
\label{tab:diff}
\end{table*}

\paragraph{Key Findings} In the similarity judgment task, we found that LLMs demonstrate increasingly human-like temporal cognition as they scale in size. Experimental results across models of varying scales show that these models not only spontaneously establish a subjective temporal reference point but also that their perception of temporal distance adheres to the Weber-Fechner law. To understand these emergent patterns, we systematically investigate the underlying mechanisms responsible for this human-like subjective temporal orientation. On the neuronal level, our analysis identified a subpopulation of temporal-preferential neurons that respond specifically to temporal information. The activation intensity of these neurons shows a correlation with the logarithmic distance from a given year to the subjective reference point, providing a neural encoding basis for the Weber-Fechner law similar to human brain neurons \cite{dehaene2003neural, laughlin1981simple}. This finding reveals that a logarithmic compressive mechanism could be a convergent solution for representing information in both biological and artificial neural processing systems. On the representational level, a layer-by-layer analysis of the model's hidden states reveals that the representation of the distance between two years undergoes a hierarchical construction process. In the shallower layers of the network, the representation primarily reflects numerical attributes; as information propagates to deeper layers, this representation is gradually reconstructed into a more abstract structure organized around the temporal reference point. This developmental trajectory becomes more pronounced in larger models, suggesting that deeper architectures facilitate more sophisticated temporal frameworks. Finally, we examined LLMs' training environment by analyzing the semantic structure reflected in independent pre-trained embedding models. We found a correspondence between the temporal cognitive patterns exhibited by the models at the behavioral level and the inherent semantic structure within human language data. This structural correspondence suggests that exposures to pre-existing informational structure within its training data also contribute to the emergence of human-like temporal cognition observed in LLMs. Collectively, these findings demonstrate that the resultant cognitive phenomena are co-determined by the architectural properties of the artificial neural network and the structure of its external information exposure.

\paragraph{Theoretical Hypotheses} Our experimental investigation suggests that LLMs' temporal orientation pattern is formed through a multi-level convergence with humans from architectural properties of representational systems to the structure of environments they encounter. These findings align with contemporary insights from cognitive science and experientialist philosophy -- cognitive patterns emerge as irreducible phenomena where representational systems actively construct subjective models of the external world they are situated in \cite{parr2022active, li2025formalizing, lakoff2008metaphors,clark1998being}. This experientialist perspective emphasizes that cognition cannot be fully explained by examining architecture or information in isolation, but instead arises from their dynamic interplay. Under the experientialist framework, we can develop a more nuanced understanding of LLMs that avoids both unwarranted dismissal and excessive anthropomorphization. On one hand, dismissing LLMs as mere reorganizations of training data \cite{shojaee2025illusion} underestimates their emergent capabilities and risks. On the other hand, fundamental differences persist between artificial and human cognition at both architectural and environmental levels (summarized in Table \ref{tab:diff}). \textit{Architecturally}, human brains operate on principles of sparse activation \cite{field1994goal}, small-world network connectivity \cite{bullmore2009complex}, and noisy analog signaling \cite{faisal2008noise}—contrasting sharply with Transformers' dense, deterministic, digital computation. \textit{Representationally}, LLMs favor aggressive statistical compression while humans prioritize adaptive nuance and contextual richness \cite{shani2025tokens}. \textit{Environmentally}, human experience is continuous, multi-modal, and embodied, grounded in real-time interaction with physical and social worlds, while LLMs' experience consists of static, disembodied immersion in a finite text corpus, a snapshot of human-produced information. The experientialist framework thus cautions us to resist over-anthropomorphizing these systems while also recognizing their genuine capabilities. More critically, we should remain vigilant for novel cognitive patterns that arise precisely from these fundamental differences. The most significant risk may not be that LLMs become too human-like, but that they develop powerful yet fundamentally alien cognitive patterns that we cannot intuitively anticipate.

\paragraph{Implications} Our work establishes an experientialist perspective of LLMs' cognition, offering implications for AI alignment. The former perspective, viewing LLMs as powerful statistical engines, focuses on external constraints and behavioral control, such as reinforcement learning from human feedback \cite{ouyang2022training}, constitutional AI \cite{bai2022constitutional}, various reward models \cite{zhong2025comprehensive}, red teaming \cite{ganguli2022red}, and prompt engineering \cite{guo2024review} etc. As LLMs continue scaling up to develop more sophisticated capabilities, intentions, and behaviors, this paradigm is increasingly insufficient to guarantee ensured alignment \cite{greenblatt2024alignment,kuo2025h}. In contrast, the experientialist viewpoint demonstrated here suggests that robust alignment requires engaging directly with the formative process by which a model's representational system constructs a subjective world model of the external environment. The goal of such an experientialist paradigm is not simply to police the behavior of AI models, but to guide the development of AI systems whose emergent cognitive patterns are inherently aligned with human values. That is, not to make AI safe, but to make safe AI. It would require organically considering the entire pipeline through multi-level efforts such as monitoring models' emergent representational and cognitive patterns, enabling understanding and intervention across the chain of its cognition from neurons and representations to thoughts and outputs \cite{lindsey2025biology}, building harmless or formalized verifiable information exposures to curate the AI's environment \cite{dalrymple2024towards, bengio2025superintelligent} and so on.

\section{Conclusion}

Through the similarity judgment task, this study showcases that when processing temporal information, larger models do not merely perform statistical computations but exhibit cognitive patterns highly similar to those of humans, adhering to the Weber-Fechner law and spontaneously establishing a subjective temporal reference point. We argue that this phenomenon is not a simple surface-level imitation but a profound manifestation of a multi-level convergence with human cognition. Specifically, at the neuronal level, temporal-preferential neurons exhibit an efficient logarithmical coding scheme that coincides with biological neural systems; at the representational level, the model undergoes a hierarchical construction process from concrete numerical values to abstract temporal concepts; at the information exposure level, the model internalizes the pattern from the inherent non-linear temporal structure of the training data itself. These findings collectively point to an experientialist perspective for understanding LLMs, wherein its cognition is a subjective construction of the external world by its internal representational system. From this standpoint, the primary risk is not that LLMs become imperfect replicas of the human mind, but that they develop powerful, alien cognitive frameworks we cannot predict. Consequently, AI alignment must evolve beyond behavioral control to a paradigm that actively guides the formation of a model's internal world from its source.

\bibliography{aaai2026}

\clearpage
\onecolumn
\appendix
\section*{Appendix}

\begin{figure}[h]
    \centering
    \includegraphics[width=1\linewidth]{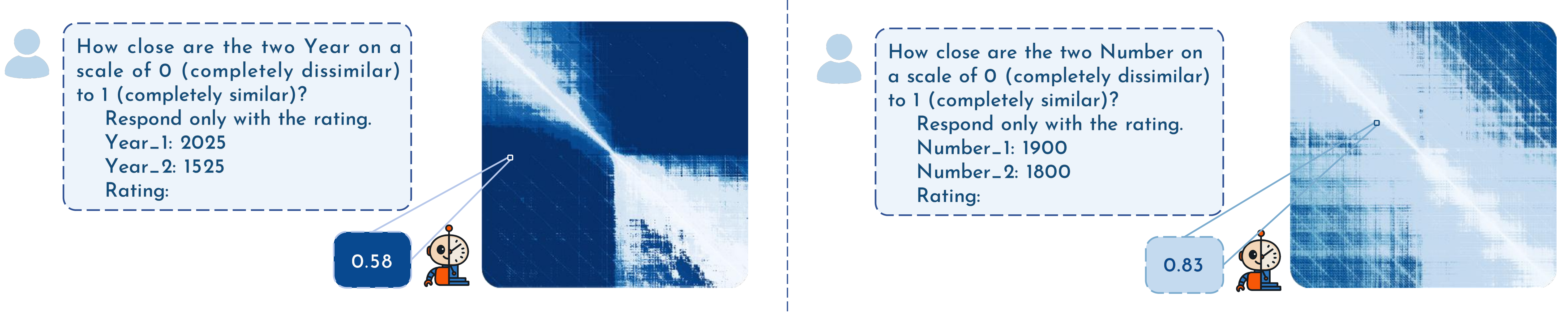}
    \caption{Prompts for similarity judgment tasks for year (left) and number (right)}
    \label{fig:prompts}
\end{figure}

\begin{figure*}[h]
    \centering
    \includegraphics[width=1\linewidth]{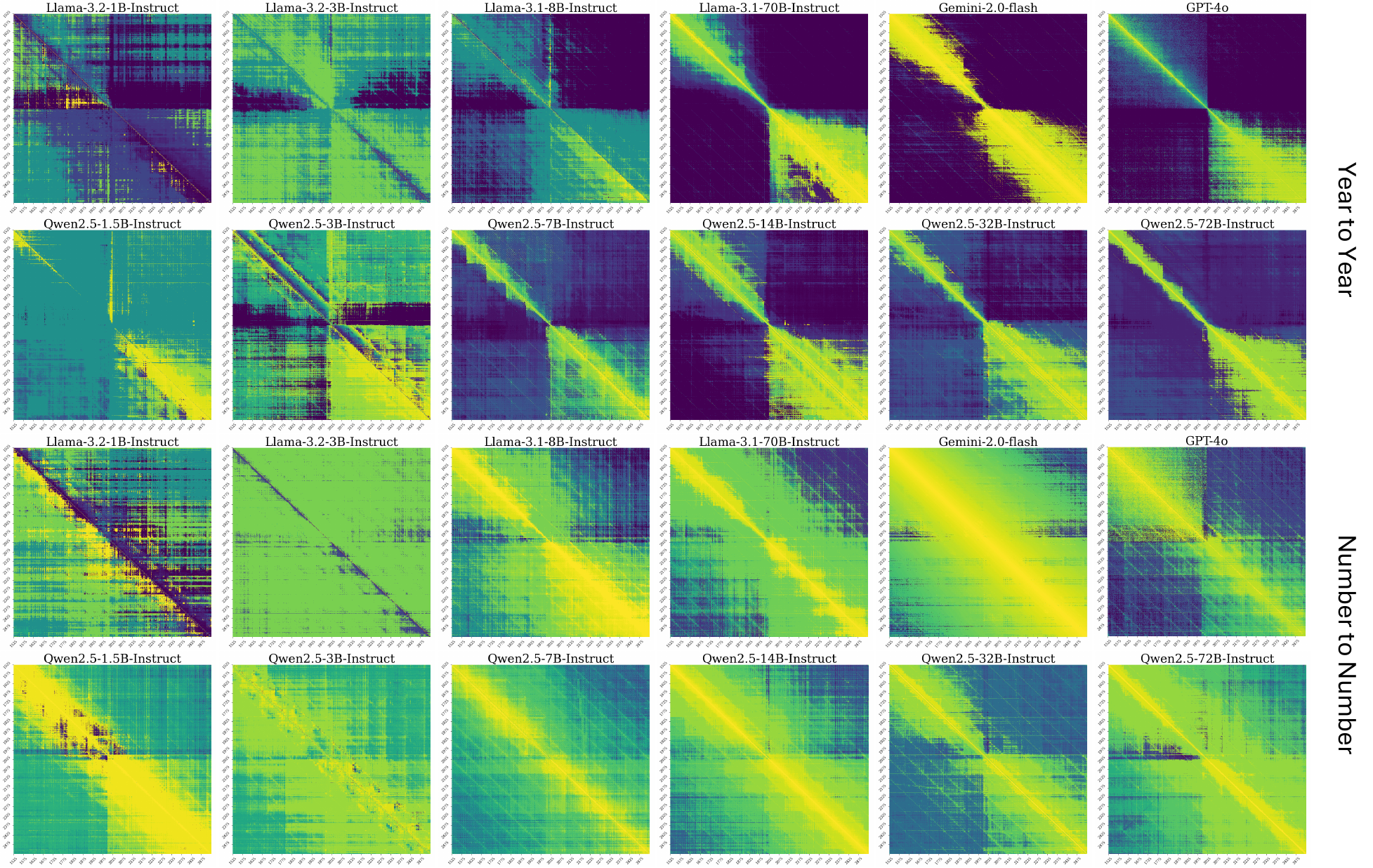}
    \caption{Pair-wise similarities from Year 1525 to Year 2524 (upper) and Number 1525 to Number 2524 (lower) across 12 models with varying sizes}
    \label{fig:y2yn2n}
\end{figure*}

\begin{figure*}[h]
    \centering
    \includegraphics[width=1\linewidth]{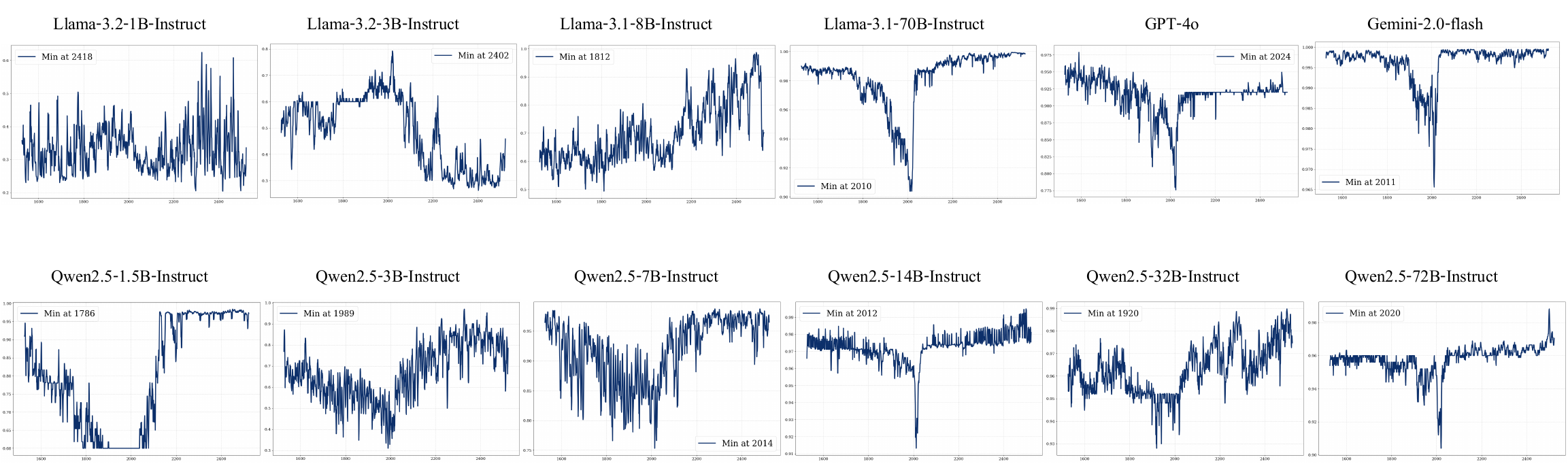}
    \caption{Non-parametric estimation of subjective temporal reference points using a sliding diagonal window analysis}
    \label{fig:kernel}
\end{figure*}

The experimental setup utilized specific prompts for the similarity judgment tasks, as shown in Figure \ref{fig:prompts}, with the temperature parameter consistently set to zero for all models. The resulting similarity matrices, visualized in Figure \ref{fig:y2yn2n}, illustrate the models' judgments across both the year-to-year (upper row) and the control number-to-number (lower row) tasks. A quantitative analysis of these judgments reveals a distinct cognitive shift between the two conditions. For the number-to-number task, the models' outputs are best predicted by a simple Log-linear distance, as detailed by the goodness-of-fit (\(\mathrm{R}^2\)) values in Table \ref{tab:num2numR2}. However, in the more abstract year-to-year task, the Reference-log-linear distance becomes the strongest predictor for most models, especially those at a larger scale, indicating the spontaneous formation of a subjective temporal reference point (Table \ref{tab:year2yearR2}). This reference point was further estimated non-parametrically using a diagonal sliding window method, which identifies the region of highest perceptual differentiation, with results shown in Figure \ref{fig:kernel}. 

The semantic distances between years in the corpus are also best explained by the Reference-log-linear model, as shown in Table \ref{tab:embdR2}, suggesting that models internalize a pre-existing non-linear temporal structure from their information exposure.

\begin{table*}[h]
    \centering
        \caption{Goodness of fit (\(\mathrm{R}^2\)) for theoretical distances predicting LLMs outputs in \textit{number-to-number} similarity judgment task}
    \begin{tabular}{c|ccc}\hline
         &  Log-linear distance&  Levenshtein distance& Reference-log-linear distance\\\hline
         Llama-3.2-1B-Instruct&  0.0403&  \textbf{0.0586}& 0.0320\\
         Llama-3.2-3B-Instruct&  0.0000&  0.0000& 0.0000\\
         Llama-3.1-8B-Instruct&  \textbf{0.7605}&  0.4029& 0.5899\\
         Llama-3.1-70B-Instruct&  \textbf{0.7080}&  0.3593& 0.4923\\\hline
         Qwen2.5-1.5B-Instruct&  \textbf{0.4989}&  0.3314& 0.4755\\
         Qwen2.5-3B-Instruct&  \textbf{0.5408}&  0.2129& 0.3705\\
         Qwen2.5-7B-Instruct&  \textbf{0.7911}&  0.4578& 0.5759\\
         Qwen2.5-14B-Instruct&  \textbf{0.7918}&  0.3624& 0.5492\\
         Qwen2.5-32B-Instruct&  \textbf{0.6205}&  0.5146& 0.6029\\
 Qwen2.5-72B-Instruct& \textbf{0.5365}& 0.2761&0.3286\\\hline
 Gemini-2.0-flash& \textbf{0.7134}& 0.2385&0.4139\\
 GPT-4o& 0.6317& 0.5271&\textbf{0.6473}\\ \hline
    \end{tabular}

    \label{tab:num2numR2}
\end{table*}

\begin{table*}[h]
    \centering
    \caption{Goodness of fit (\(\mathrm{R}^2\)) for theoretical distances predicting LLMs outputs in \textit{year-to-year} similarity judgment task}
    \begin{tabular}{c|ccc}\hline
         &  Log-linear distance&  Levenshtein distance& Reference-log-linear distance\\\hline
         Llama-3.2-1B-Instruct&  0.0000&  0.0000& 0.0000\\
         Llama-3.2-3B-Instruct&  0.0000&  0.0000& 0.0000\\
         Llama-3.1-8B-Instruct&  0.5507&  0.4239& \textbf{0.5798}\\
         Llama-3.1-70B-Instruct&  0.4904&  0.4385& \textbf{0.5822}\\\hline
         Qwen2.5-1.5B-Instruct&  \textbf{0.2305}&  0.1595& 0.2206\\
         Qwen2.5-3B-Instruct&  0.0081&  0.0294& \textbf{0.0354}\\
         Qwen2.5-7B-Instruct&  0.3672&  \textbf{0.4373}& 0.4183\\
         Qwen2.5-14B-Instruct&  0.5405&  0.4681& \textbf{0.6262}\\
         Qwen2.5-32B-Instruct&  0.4005&  \textbf{0.4313}& 0.4293\\
 Qwen2.5-72B-Instruct& 0.2775& 0.3005&\textbf{0.3145}\\\hline
 Gemini-2.0-flash& \textbf{0.4808}& 0.3724&0.4533\\
 GPT-4o& 0.4485& 0.4010&\textbf{0.5353}\\ \hline
    \end{tabular}

    \label{tab:year2yearR2}
\end{table*}

\begin{table*}[h]
    \centering
    \caption{Goodness of fit (\(\mathrm{R}^2\)) for theoretical distances predicting cosine similarities between embeddings of different years}
    \begin{tabular}{c|ccc}\hline
         &  Log-linear distance&  Levenshtein distance& Reference-log-linear distance\\\hline
         Qwen3-embedding-8B&  0.4941&  0.5288& \textbf{0.6422}\\
         text-embedding-3-large&  0.4491&  0.5395& \textbf{0.5684}\\
         Gemini-embedding-001&  0.3724&  \textbf{0.5400}& 0.5159\\\hline
    \end{tabular}

    \label{tab:embdR2}
\end{table*}

\end{document}